\documentclass[10pt,twocolumn,letterpaper]{article}
\usepackage[final]{cvpr}
\usepackage{times}
\usepackage{epsfig}
\usepackage{graphicx}
\usepackage{float}
\usepackage{amsmath}
\usepackage{amssymb}
\usepackage{soul}
\usepackage{multicol}
\usepackage{blindtext}
\usepackage{here}
\usepackage{caption}
\usepackage{makecell}
\usepackage[normalem]{ulem}
\usepackage{multirow}
\usepackage{booktabs}
\usepackage{color, colortbl}
\usepackage{pifont}
\definecolor{Gray}{gray}{0.9}
\usepackage[pagebackref=true,breaklinks=true,letterpaper=true,colorlinks,bookmarks=false]{hyperref}

\definecolor{LightCyan}{rgb}{0.88,1,1}
\definecolor{LightYellow}{rgb}{1,0.98,0.8}
\definecolor{lightgray}{rgb}{0.9,0.9,0.9}

\newcommand{\ourTitle}{Capturing and Inferring Dense Full-Body Human-Scene Contact}

\newcommand{\supmat}{Sup.~Mat.\xspace}

\newcommand{\smpl}{\mbox{SMPL}\xspace}

\newcommand{\smplX}{\mbox{SMPL-X}\xspace}

\newcommand{\alphapose}{\mbox{AlphaPose}\xspace}
\newcommand{\mvpose}{\mbox{MvPose}\xspace}

\newcommand{\mocap}{\mbox{MoCap}\xspace}
\newcommand{\humor}{\mbox{HuMoR}\xspace}

\newcommand{\sota}{\mbox{SOTA}\xspace}
\newcommand{\twoD}{2D\xspace}
\newcommand{\threeD}{3D\xspace}

\newcommand{\na}{\mbox{N/A}\xspace}
\newcommand{\hps}{\mbox{HPS}\xspace}
\newcommand{\hsc}{\mbox{HSC}\xspace}
\newcommand{\hsi}{\mbox{HSI}\xspace}
\newcommand{\hoi}{\mbox{HOI}\xspace}
\newcommand{\multiv}{\mbox{multiview}\xspace}
\newcommand{\websiteURL}{\url{https://rich.is.tue.mpg.de}}

\renewcommand{\etal}{et al.\xspace}
\renewcommand{\ie}{i.e.\xspace}
\renewcommand{\eg}{e.g.\xspace}

\newcommand{\numV}{$C$\xspace}
 
\newcommand{\contactvector}{\mathbf{c}}

\newcommand{\numVert}{V\xspace} 

\newcommand{\methodname}{BSTRO\xspace} 
\newcommand{\datasetname}{RICH\xspace} 
\newcommand{\numimage}{577K\xspace} 
\newcommand{\numpose}{90K\xspace} 
\newcommand{\numscene}{5\xspace}
\newcommand{\nummvvideo}{142\xspace} 
\newcommand{\nummvvideoTr}{62\xspace} 
\newcommand{\nummvvideoTe}{52\xspace} 
\newcommand{\nummvvideoVal}{28\xspace} 
\newcommand{\numsub}{22\xspace}

\newcommand{\cmark}{\ding{51}}%
\newcommand{\xmark}{\ding{55}}%

\def\cvprPaperID{933} 
\def\confYear{2022}
\def\confName{}

\begin{document}

\title{\ourTitle}

\author{
Chun-Hao P.~Huang\textsuperscript{1}\quad
Hongwei Yi\textsuperscript{1}\quad
Markus H{\"o}schle\textsuperscript{1}\quad
Matvey Safroshkin\textsuperscript{1}\quad
Tsvetelina Alexiadis\textsuperscript{1}\quad \\
Senya Polikovsky\textsuperscript{1}\quad 
Daniel Scharstein\textsuperscript{2}\quad
Michael J.~Black\textsuperscript{1}\\
\textsuperscript{1}Max Planck Institute for Intelligent Systems, T{\"u}bingen, Germany
\quad
\textsuperscript{2}Middlebury College
\\
{\tt\small \{paul.huang, firstname.lastname, black\}@tuebingen.mpg.de, schar@middlebury.edu}\\
}

\twocolumn[{
    \renewcommand\twocolumn[1][]{#1}
    \maketitle
    \centering
    \vspace{-0.2em}
    \begin{minipage}{1.00\textwidth}
    \centering
        \includegraphics[trim=000mm 000mm 000mm 000mm, clip=false, width=1.00 \linewidth]{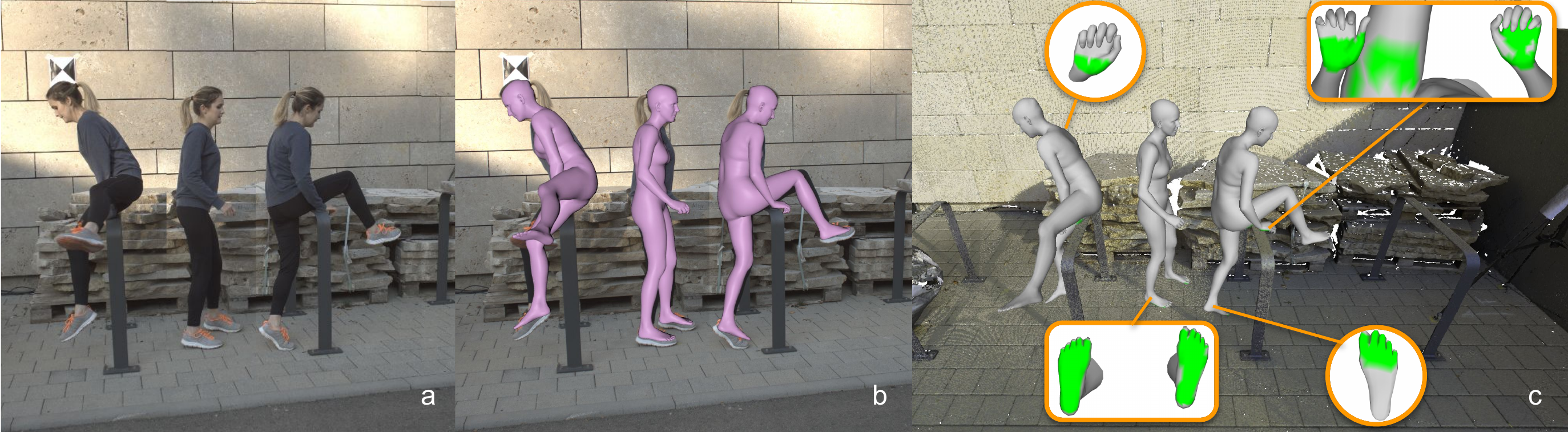}
    \end{minipage}

    \captionof{figure}{{\bf \datasetname} is a new dataset containing videos of people in natural scenarios and standard clothing together with ground-truth 3D body pose and shape (a-b). 
   A key novelty of \datasetname is that it also contains 3D scene scans, which enable dense and accurate labeling of human-scene contact (c, green). 
   We exploit this to learn a regressor called \methodname that takes an image and infers human-scene contact.
    }\label{fig:teaser}
    \vspace*{+01.50em}
}]


\maketitle


\begin{abstract}
Inferring human-scene contact (\hsc) is the first step toward understanding how humans interact with their surroundings.
While detecting 2D human-object interaction (\hoi) and reconstructing 3D human pose and shape (\hps) have enjoyed significant progress,
reasoning about 3D human-scene contact from a single image is still challenging.
Existing \hsc detection methods consider only a few types of predefined  contact, 
often reduce the body and scene to a small number of primitives,
and even overlook image evidence.
%
To predict human-scene contact from a single image, we address the limitations above from both data and algorithmic perspectives. 
We capture a new dataset called \datasetname for ``Real scenes, Interaction, Contact and Humans.''
\datasetname contains multiview outdoor/indoor video sequences at 4K resolution, ground-truth 3D human bodies captured using markerless motion capture, 3D body scans, and high resolution 3D scene scans.
A key feature of \datasetname is that it also contains accurate vertex-level contact labels on the body. 
%
Using \datasetname, we  train a network that predicts dense body-scene contacts from a single RGB image.
Our key insight is that regions in contact are always occluded so the network needs the ability to explore the whole image for evidence. 
We use a transformer to learn such non-local relationships and propose a new Body-Scene contact TRansfOrmer (\methodname).
Very few methods explore 3D contact; those that do
focus on the feet only, detect foot contact as a post-processing step, or infer contact from body pose without looking at the scene.
To our knowledge, \methodname is the first method to directly estimate 3D body-scene contact from a single image.
We demonstrate
that \methodname significantly outperforms the prior art. 
Our code and dataset are available for research purposes at: \websiteURL

\end{abstract}
\section{Introduction}
\label{introduction}

Understanding human actions and behaviors has long been studied in computer vision, with 
applications in robotics, healthcare, virtual try-on, AR/VR, and beyond. 
Remarkable progress has been made in both \twoD human pose detection \cite{cao2019openpose,hidalgo2019singlenetwork, jin2020whole,li2021human, sun2019deep, xiao2018simple} and \threeD human pose and shape estimation (HPS) from a single image \cite{Bogo:ECCV:2016,kanazawa2018end,Kocabas_PARE_2021,kolotouros2019spin,lin2021end-to-end,SMPL-X:2019,xiang2019monocular,zhangoohcvpr20}, thanks to realistic datasets annotated with \twoD keypoints \cite{andriluka14benchmark,johnson2011learning,lin2014microsoft} and 3D data \cite{h36m_pami,joo2021eft,mono-3dhp2017,sigal2010humaneva,vonMarcard20183dpw}. 
Despite this progress, something important is missing.
Even the most basic human activities, such as walking, involve interaction with the surrounding environment.
Fundamentally, human-scene interaction (\hsi) involves the contact relationships between a \threeD human and a \threeD scene, \ie, human-scene contact (\hsc).
Existing \hps methods, however, largely ignore the scene and estimate human poses and/or shapes in isolation,  
often leading to physically implausible results.

Since reconstructing the full \threeD scene from a single image is challenging, 
recent \hps methods tackle this problem by making several simplifying assumptions about the scene and/or body. 
Many methods consider only the contact between feet and ground \cite{RempeContactDynamics2020,Shimada2020PhysCap,Xie_2021_ICCV,yuan2021simpoe,zanfir2018monocular,Zhang_2021_ICCV,zou2020reducing}, or assume the ground is a even plane \cite{rempe2021humor}, which is often violated, \eg, walking up stairs. 
To infer contact, many state-of-the-art (\sota) methods use \mocap datasets \cite{AMASS:2019,mixamo} to train a contact detector \cite{RempeContactDynamics2020,Zhang_2021_ICCV,zou2020reducing}.
Others exploit physics simulation \cite{Shimada2020PhysCap,yuan2021simpoe} or physics-inspired objectives \cite{Xie_2021_ICCV} but reduce the body representation to a small set of primitives.
Surprisingly, none of these methods use image evidence when predicting human-scene contact.
This is primarily due to the lack of datasets with images and \threeD contact ground truth.

Many methods do estimate human object interaction (HOI) from images but constrain the reasoning to \twoD image regions \cite{kim2021hotr,zou2021_hoitrans,qi2018learning,xu2019learning,wang2019deep}.
That is, they estimate bounding boxes or heatmaps in the image corresponding to contact but do not relate these to the \threeD body.

In this work, 
we address this problem with
a framework that estimates 3D contact on the body directly from a single image.
We make two main contributions.
First, we create a new dataset that accurately 
captures human-scene contact by extending a markerless \mocap method to markerless \hsc capture.
Specifically, we capture multiview video sequences at 4K resolution in both indoor and outdoor environments. 
We also capture the precise 3D geometry of the scene using a laser scanner. 
Additionally, we capture high-resolution 3D scans of all subjects in minimal clothing and fit the SMPL-X body model \cite{SMPL-X:2019} to the scans.
Our markerless \hsc approach allows us to compute accurate per-vertex scene contact, as visualized in Fig.~\ref{fig:teaser}c.

Compared to the PROX dataset \cite{PROX:2019}, which captures \hsc with monocular RGB-D input, multiview data has two advantages: 
(1) it effectively resolves occlusions, leading to better reconstructed bodies and consequently more accurate scene contact;
(2) it works for outdoor environments, as shown in Fig.~\ref{fig:teaser}.

The resulting dataset, called \datasetname (``Real scenes, Interaction, Contact and Humans''), provides: 
(1) high-resolution multiview images of single or multiple subjects interacting with a scanned 3D scene, 
(2) dense full-body scene-contact labels, 
(3) high-quality outdoor/indoor scene scans,
(4) high-quality 3D human shapes and poses, and
(5) dynamic backgrounds and moving cameras. 

To estimate vertex-level \hsc from a single color image, we develop \methodname (Body-Scene contact TRansfOrmer), and train it with \datasetname.
Our key insight in building \methodname is that contact is not directly observable in images due to occlusion; 
thus, to infer contact, the network architecture must be able to explore the whole image for evidence.
The transformer architecture enables \methodname 
to learn non-local relationships
and use scene information to ``hallucinate'' unobserved contact.
We employ a multi-layer transformer \cite{vaswani_transformers},
which has been successfully employed for
natural-language processing \cite{devlin2018bert} and \hps estimation with occlusion \cite{lin2021end-to-end}.

In summary, our key contributions are:
(1) We present \datasetname, a novel dataset that captures people interacting with complex scenes. 
It is the first dataset that provides \emph{both scans of outdoor scenes and images} for monocular \hsc estimation, unlike existing methods \cite{Guzov2021cvpr,PROX:2019}, which lack one or the other.
(2) We propose \methodname, a monocular \hsc detector.
It is \emph{body-centric} so it does not require \threeD scene reconstructions to infer contact. 
Unlike POSA \cite{Hassan:CVPR:2021}, which is also body-centric, \methodname directly estimates dense scene contact from the input image without reconstructing bodies.
(3) We evaluate recent \hsc methods and show that \methodname gives \sota results.
(4) Since \datasetname has pseudo-ground-truth body fits, we also evaluate \sota \hps methods and analyze their performance with respect to scene-contact,
which is not supported by existing \hps datasets \cite{h36m_pami,Patel:CVPR:2021,vonMarcard20183dpw}.
We confirm that the performance of a \sota \hps method \cite{feng2012pixie} 
degrades in
the presence of scene contact.

\begin{table}[t]
\centering
\footnotesize
\begin{tabular}{l|c|c|c}
\toprule
    \multirow{2}{*}{\textbf{Methods}} & \textbf{Scene Contact} & \multirow{2}{*}{\textbf{Body}} & \textbf{Contact Cues}   \\ 

      & Body / Scene &  &  Train / Test  \\ 
     \hline
     Zanfir \etal~\cite{zanfir2018monocular}& \multirow{4}{*}{\shortstack[c]{foot joints \\ / ground}}  & mesh & - / dist. \\ 
     \cline{1-1} \cline{3-4}
     Zou \etal~\cite{zou2020reducing} & & \multirow{2}{*}{joint} & both 2D vel.\\
     \cline{1-1} \cline{4-4}
     Rempe \etal~\cite{RempeContactDynamics2020} &  &  & \multirow{4}{*}{\shortstack[c]{vel. \& dist. \\ / vel. \& dist.}} \\ 
     \cline{1-1} \cline{3-3}
     PhysCap \cite{Shimada2020PhysCap}  &  & part &  \\ 
     \cline{1-2} \cline{3-3}
     \humor \cite{rempe2021humor} & 8 joints / ground & \multirow{2}{*}{\shortstack[c]{mesh}} &   \\ 
     \cline{1-2}
     LEMO \cite{Zhang_2021_ICCV} & foot vert.~/ ground &  &  \\ 
     \hline
     SimPoE \cite{yuan2021simpoe}  & \multirow{2}{*}{\shortstack[c]{foot parts \\ / ground}} & \multirow{2}{*}{part} & \multirow{2}{*}{\shortstack[c]{physics \\simulation}} \\ 
     \cline{1-1}
     Xie \etal~\cite{Xie_2021_ICCV} &  &  &  \\ 
     \hline
     HolisticMesh \cite{weng2020holistic}  &  \multirow{6}{*}{\shortstack[c]{dense body mesh \\ / scene mesh}} &  \multirow{6}{*}{mesh} & - / dist. \\ 
     \cline{1-1} \cline{4-4}
     PROX \cite{PROX:2019} &   &  & - / dist. \\ 
     \cline{1-1} \cline{4-4}
     PHOSA \cite{zhang2020phosa} &  &  & - / dist.\\
     \cline{1-1} \cline{4-4}
     Zhang \etal~\cite{zhang2019psi} &  &  & dist.~/ -\\
     \cline{1-1} \cline{4-4}
     PLACE \cite{zhang2020place} &  &  & dist. \\
     \cline{1-1} \cline{4-4}
     POSA \cite{Hassan:CVPR:2021}  &  &  & dist.~/ pose\\ 
     \hline
     \rowcolor{lightgray}
     \methodname (ours) & as above  & mesh & dist.~/ image  \\ 
     \hline 
     \hline
     \textbf{Datasets} & \textbf{Contact Label} & \textbf{Img} &  \textbf{Scene}   \\ 
     \hline
     MTP \cite{Mueller:CVPR:2021} & self-contact & \cmark & \na \\ 
     \hline
     GRAB \cite{taheri2020grab} & hand-object & \xmark & \na \\ 
     \hline
     ContactHands \cite{Narasimhaswamy2020nips} & hand-X$^\ddagger$ & \cmark & \na\\
     \hline
     Fieraru \etal~\cite{Fieraru_2020_CVPR} & person-person & \cmark & \na\\
     \hline
     Fieraru \etal~\cite{Fieraru_2021_AAAI} & self-contact & \cmark & \na\\
     \hline
     PiGraph \cite{savva2016pigraphs} & joint-scene & \cmark & RGBD scans\\
     \hline
     i3DB \cite{iMapper2018} & \na & \cmark & CAD \\
     \hline
     GPA \cite{wang2019geometric} & \na & \cmark & Cubes \\ 
     \hline
     Guzov \etal~\cite{Guzov2021cvpr} & foot-ground & \cmark$^\mathparagraph$ & laser scans \\
     \hline
     PROX \cite{PROX:2019} & body-scene  & \cmark & RGBD scans \\ 
     \hline
     \rowcolor{lightgray}
     \datasetname (ours) & body-scene  & \cmark &  laser scans \\  
    
    
    \bottomrule
\end{tabular}
\caption{\textbf{Comparison of contact-related methods and datasets}. $^\ddagger$: X can be self, person and object. 
$^\mathparagraph$: egocentric images. Vert.: vertex; vel.: velocity; dist.: distance.}
\label{table:relatedwork}
\vspace{-3ex}
\end{table}
\section{Related Work}\label{related_work}
We review existing methods that consider contact between humans and scenes. 
Since many of them employ a 3D body reconstruction method as a backbone in the pipeline, 
we first briefly discuss recent \hps trends and then focus on how the prior art incorporates scene contact.

\subsection{Human Pose and Shape Estimation (\hps)}
\noindent\textbf{Monocular \hps} methods reconstruct \threeD human bodies from a single color image. 
Many methods output the parameters of statistical 3D body models~\cite{Anguelov2005scape,joo2018totalcapture,SMPL:2015,SMPL-X:2019,xu2020ghum}.
%
SMPLify \cite{Bogo:ECCV:2016} fits the SMPL model to the output of a \twoD keypoint detector \cite{Leonid2016DeepCut} and we build on it here.

In contrast, deep neural networks regress body-model parameters directly from pixels~\cite{ExPose:2020,Dwivedi_DSR_2021,feng2012pixie,guler_2019_CVPR,kanazawa2018end,kocabas2019vibe,Kocabas_PARE_2021,kolotouros2019spin,rong2021frankmocap,ROMP:ICCV:2021,BEV}. 
To deal with the lack of in-the-wild 3D ground truth, 
some methods use \twoD keypoints~\cite{kanazawa2018end,Tan,tung2017self} or linguistic attributes \cite{Shapy:CVPR:2022} as weak supervision, 
while some directly fine-tune the network w.r.t.~an input image at test time \cite{joo2021eft}. 
Kolotouros \etal~\cite{kolotouros2019spin} combine HMR~\cite{kanazawa2018end} and SMPLify~\cite{Bogo:ECCV:2016} in a training loop for better 3D supervision. 
%
On the other hand, non-parametric or model-free approaches
directly estimate 3D vertex locations without body parameters \cite{Choi_2020_ECCV_Pose2Mesh,kolotouros2019cmr,lin2021end-to-end,lin2021-mesh-graphormer,Moon_2020_ECCV_I2L-MeshNet,dittadi2021full,Zeng_2020_CVPR}.
We refer readers to \cite{tian2022hmrsurvey,zheng2022deep} for a comprehensive review.
None of the above methods estimate \hsc. 

\noindent\textbf{Markerless \mocap} exploits synchronized videos from multiple calibrated cameras and has a long history with commercial solutions, but these focus on estimating a 3D skeleton.
To model \hsc, we need to extract a full 3D body shape and, therefore, focus on such methods here.
Early methods, either bottom-up \cite{belagiannis20143d,grauman2003inferring,sigal2008combined}  or top-down \cite{Balan:CVPR:2007,gall2009motion,vlasic2008articulated}, 
are fragile, need subject-specific templates and manual input, and do not generalize well to in-the-wild images.

Powered by CNNs, recent methods leverage multiview consistency to improve keypoint detection \cite{epipolartransformers2020cvpr,iskakov2019learnable,Qiu_2019_ICCV,to2020voxelpose}, 
to re-identify subjects across views \cite{Dong_2019_CVPR} or across view and time \cite{Dong_2021_PAMI,Zhang_2020_CVPR}, but they estimate only joints, not body meshes. 
Dong \etal~\cite{dong2021shape} reconstruct SMPL bodies for multiple subjects and Zhang \etal~\cite{lightcap2021} additionally estimate hands and facial expressions. 
\textcolor{black}{They demonstrate results for lab scenarios, while our \hsc capture method in Sec.~\ref{sub-sect:mvinit} works in less constrained outdoor scenes}. 

All methods above reconstruct human bodies in isolation without taking into account the interaction with scenes. 
Consequently, the results often contain physically implausible artifacts such as foot skating and ground penetration.

\subsection{Human Scene Interaction (\hsi)}
\noindent \textbf{\twoD Human-Object Interaction (\hoi)} methods localize \twoD image regions with \hoi and recognize the semantic interactions in them. 
Most methods represent humans and objects very roughly as bounding boxes \cite{kim2021hotr,zou2021_hoitrans}; only a few use body meshes for humans and spheres for objects \cite{li2020detailed}. %

\noindent \textbf{\threeD Contact.} Knowing which part of the body and  scene are in contact provides compact yet rich information that enables many applications, such as \hsi recognition \cite{brahmbhatt2019contactdb} or placing virtual humans into a scene \cite{Hassan:CVPR:2021}.
The upper part of Table \ref{table:relatedwork} summarizes how body-scene contact gets incorporated in methods of different goals and tasks.

Early work uses scene contact as part of the \hsi feature \cite{iMapper2018, savva2016pigraphs} but represents a human body roughly as a stick figure.
Recent \hps methods \cite{PROX:2019,rempe2021humor,RempeContactDynamics2020,zou2020reducing} use contact to improve the estimated body poses. 
Ideally, when both the body and scene are ``perfectly reconstructed,'' 
applying a threshold to the \threeD Euclidean distances between them is sufficient to infer accurate contact.
Prior work takes this thresholding approach to annotate contact \cite{Guzov2021cvpr,PROX:2019,Mueller:CVPR:2021,taheri2020grab}.
At test time, PROX \cite{PROX:2019} assumes scene scans to be known a priori; PHOSA \cite{zhang2020phosa} estimates \threeD objects, \threeD people, and the contacts between them but only for a limited class of objects. 
Since reconstructing a \threeD scene in high quality with correct layout and spatial arrangement is still an open challenge \cite{yi2022mover}, 
monocular \hsc detection methods resort to other heuristics.
The most common one is a zero-velocity assumption; \ie, surfaces in contact should not slide relative to each other.
This assumption is widely employed to reduce foot-skating artifacts \cite{rempe2021humor,RempeContactDynamics2020,Shimada2020PhysCap,zou2020reducing}. 
Some of these detect contact with a separate neural network at test time, taking the \twoD/\threeD joints in a temporal window as input \cite{RempeContactDynamics2020,Shimada2020PhysCap,zou2020reducing}, while others integrate it in a body motion prior \cite{rempe2021humor,Zhang_2021_ICCV}. 
These approaches use \mocap datasets such as AMASS \cite{AMASS:2019} and Mixamo \cite{mixamo} to build training data, 
where contact is automatically labelled via thresholding the distance to the ground and/or the velocity.

POSA \cite{Hassan:CVPR:2021} observes that scene contact is correlated with body poses 
and introduces a generative model to sample contact given a posed mesh. 
Some methods \cite{Shimada2020PhysCap,Xie_2021_ICCV,yuan2021simpoe} apply physics 
to encourage foot-ground contact and ensure physically plausible motions. However, they have to approximate the body as a set of boxes, cylinders or spheres. 
MOVER \cite{yi2022mover} uses human scene contact to improve monocular estimation of 3D scene layout.
%
%

All these approaches first reconstruct bodies (2D or 3D), and then reason about contact, effectively ignoring valuable image information.  
To go further, we need a dataset consisting of natural images and 3D body-scene contact labels. 
As summarized in the lower part of Table~\ref{table:relatedwork}, many existing contact-related datasets consider self contact \cite{Fieraru_2021_AAAI,Mueller:CVPR:2021} or person-person contact \cite{Fieraru_2020_CVPR}, but not \hsc.
The most relevant datasets for \hsc are \cite{Guzov2021cvpr} and PROX \cite{PROX:2019}. 
The former provides egocentric images for localization, which are not suitable for \hsc detection from images.
PROX \cite{PROX:2019} can be used for our task but it consists of only indoor scenes and is of lower quality.
The ground-truth bodies in PROX are computed by fitting to RGBD data, which is sensitive to occlusions. 
This not only limits the type of \hsi in the dataset (mostly walking, sitting, lying) but also influences the quality of body fits. 

\section{Methods: \datasetname Dataset}
\label{methods}
\noindent \textbf{Overview and preliminaries}. 
Unlike \cite{RempeContactDynamics2020,Shimada2020PhysCap,Xie_2021_ICCV,Yu:2021:MovingCam}, which represent a body as a set of coarse geometry primitives, 
we follow \cite{PROX:2019,Hassan:CVPR:2021} to capture realistic human-scene contact with a parametric \smplX body model \cite{SMPL-X:2019}.  
The vertex locations on a SMPL-X mesh $M(\theta,\beta,\psi) \subset \mathbb{R}^3$ are controlled by parameters for pose $\theta$, shape $\beta$, and facial expression $\psi$. 
$\theta$ consists of body pose $\theta_b$ and hand pose $\theta_{h}$.
Hand pose $\theta_{h}$ is a function $\theta_{h}(Z_{h})$ of a PCA latent vector $Z_{h} \in \mathbb{R}^{12}$. 

Given videos captured by \numV synchronized cameras, we first identify each subject across views and across time with \cite{Dong_2019_CVPR,xiu2018poseflow}. 
For each identified subject, we reconstruct a \smplX body by a multiview fitting method that is robust to noisy 2D keypoint detections, 
and we place it in a pre-scanned scene to compute body-scene contact (Sec.~\ref{sub-sect:mvinit}). 
With this approach, we build a monocular body-scene interaction dataset (\datasetname) comprising \numimage images paired with \smplX parameters and scene contact labels (Sec.~\ref{sub-sect:dataset}). 

\subsection{Capturing Dense Body-Scene Contact} \label{sub-sect:mvinit}
We first track subjects temporally in each video with \alphapose \cite{xiu2018poseflow}, 
followed by \mvpose \cite{Dong_2019_CVPR} to match the tracklets across views. 
Other methods that build such \emph{4D associations} \cite{Dong_2021_PAMI, Zhang_2020_CVPR} could also be applied here.

At time $t$, we now have at most \numV bounding boxes of the same person and we aim to reconstruct the body. 
To this end, we adapt SMPLify-X \cite{SMPL-X:2019} to accommodate multiview data. 
SMPLify-X optimizes the pose $\theta$, shape $\beta$ and facial expression $\psi$ of SMPL-X to match the observed 2D keypoints~\cite{cao2019openpose} by minimizing the following objective:
\begin{equation}
\begin{split}
  &E(\beta,\theta,\psi) = E_J + E_{\text{reg}} \\
  &E_{\text{reg}} = \lambda_{\theta_b} 		E_{\theta_b}  +
		\lambda_{\alpha}		 	E_{\alpha}			+ 	\lambda_{\beta}    		E_{\beta}	+ 	\lambda_{\mathcal{E}}    		E_{\mathcal{E}}			+	\lambda_{\mathcal{C}}    		E_{\mathcal{C}},
  \label{eq:objective-smplifyx}
  \end{split}{}
\end{equation}
where $E_J$ is the data term, and $E_{\text{reg}}$ includes several regularization terms: $\theta_b$ is the pose vector for the body, 
which is a function $\theta_b(Z_b)$, where $Z_b \in \mathbb{R}^{32}$ is a VAE latent representation and $E_{\theta_b}$ is an $L_2$ prior defined on $Z_b$. 
$E_{\alpha}(\theta_b)$ penalizes strong bending of elbows and knees. 
$E_{\beta}(\beta)$ is an  $L_2$ prior on the body shape and $E_{C}$ is a term penalizing mesh-intersections. 
$\lambda$'s denote weights for each respective term. Interested readers are referred to~\cite{SMPL-X:2019} for details. 

\begin{figure}[t]
  \centering
  \includegraphics[width=1\linewidth]{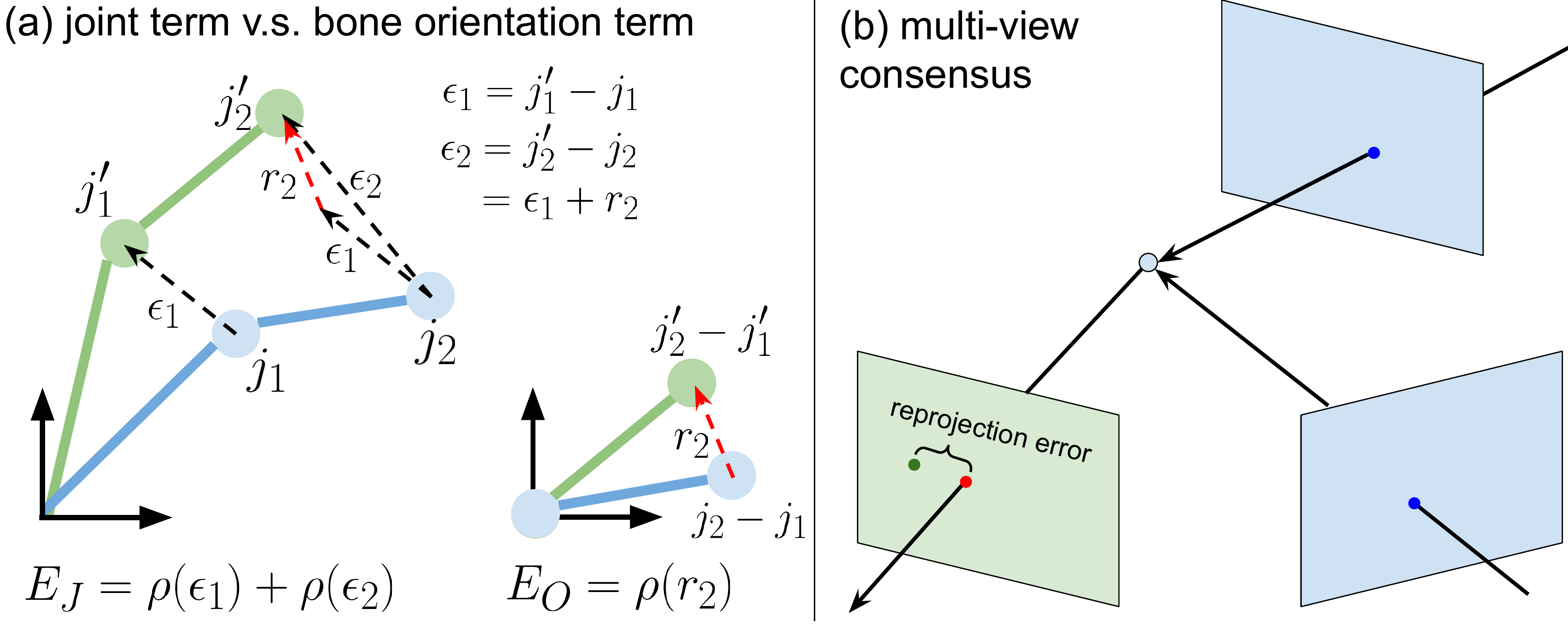}
  \caption{Illustration of bone orientation term and \multiv consensus. $\rho$ is a Geman-McClure robust estimator \cite{GemanMcClure1987}. See text and \supmat~for more discussion.}
  \label{fig:bl_mvconsensus}
\end{figure}
\noindent\textbf{Multiview per-person reconstruction.} 
For each person, we compute 2D keypoints \cite{cao2019openpose} in each camera $c$. 
Instead of fitting them using SMPLify-X in each view, we combine all 2D landmarks in a multiview energy term: $ \sum_c E_J^c $. 
Unlike in \cite{SMPL-X:2019}, where one needs to estimate camera translation first, 
the perspective projection here is well defined by the pre-calibrated intrinsics and extrinsics. 
To pursue high-quality fits, body shape $\beta$ is estimated in advance by registering a SMPL-X template to minimally-clothed 3D scans following \cite{Hirshberg:ECCV:2012}.
$\beta$ is hence no longer a free variable in Eq.~\ref{eq:objective-smplifyx} and we set $\lambda_{\beta}=0$.
In addition to $E_J$, which measures joint errors, we also use $E_O$ that measures errors in ``bone orientations.''
Figure \ref{fig:bl_mvconsensus}(a) illustrates the intuition behind this term.
Since posing human bodies requires traversing a kinematic chain, with the joint term $E_J$, 
the error of parent joints $\epsilon_1$ is accumulated in the error of child joints $\epsilon_2$.
When $\|\epsilon_2\|$ gets too large, the influence is downweighted because our robust loss treats it as an outlier. 
Instead, $E_O$ factors out the errors of ancestors and focuses on the error of the joint per se. 
Our final objective is $E_{\text{mv}}(\theta,\psi) = \sum_c E_J^c + \sum_c E_O^c + E_{\text{reg}}$.

Due to noisy \twoD detections, keypoints in each view often disagree with each other.
One may count on the robustifier to identify outliers and reduce their contribution.
This depends, however, on the current estimated body in the optimization, so it assumes good initialization.
Instead, we check the \multiv consistency of landmarks as illustrated in Fig.~\ref{fig:bl_mvconsensus}(b).
For each joint, we take the detections in two views (blue), 
triangulate a 3D point and project it to the third view (green).
If the distance between the projected point (red) and the detection (green) in the third view is large, 
that means the three detections do not agree with each other and at least one of them is wrong.
Instead of making hard decision separating outliers from inliers, 
we exhaustively compute all triplets of views, 
accumulate the reprojection error and downweight the contribution in $\sum_c E_J^c$ for views with high errors.
We term this \emph{\multiv consensus}, as it behaves like a soft majority voting mechanism.
As long as there are more correct detections  than  wrong ones, 
it can reduce the influence of noisy landmarks, independent of the current body estimate.

To further avoid local minima, we apply a state-of-the-art in-the-wild body regressor (PARE \cite{Kocabas_PARE_2021}) to initialize $\theta$. 
We run PARE on the bounding box from each view, 
fuse the results by averaging the poses, 
and covert the fused body from SMPL to SMPL-X.
The SMPL-X body pose gives the initial value of $\theta$ for minimizing $E_{\text{mv}}$. 
We first solve $E_{\text{mv}}$ 
in a frame-wise manner
and then refine a batch of $T$ frames
jointly with two additional terms on body and hand motions:
$E_{\text{batch}}(\theta_1, \cdots, \theta_T) = \sum_{t=1}^T E_{\text{mv}}^t + \lambda_\text{sm}^b E_{\text{sm}}^b+ \lambda_\text{sm}^h E_{\text{sm}}^h $. $E_{\text{sm}}^b$ is the smoothness term in \cite{Zhang_2021_ICCV} and $E_{\text{sm}}^h$ encourages neighboring frames to have similar hand-pose PCA vectors $Z_h$.

\begin{figure}
  \centering
  \includegraphics[clip,width=1\linewidth]{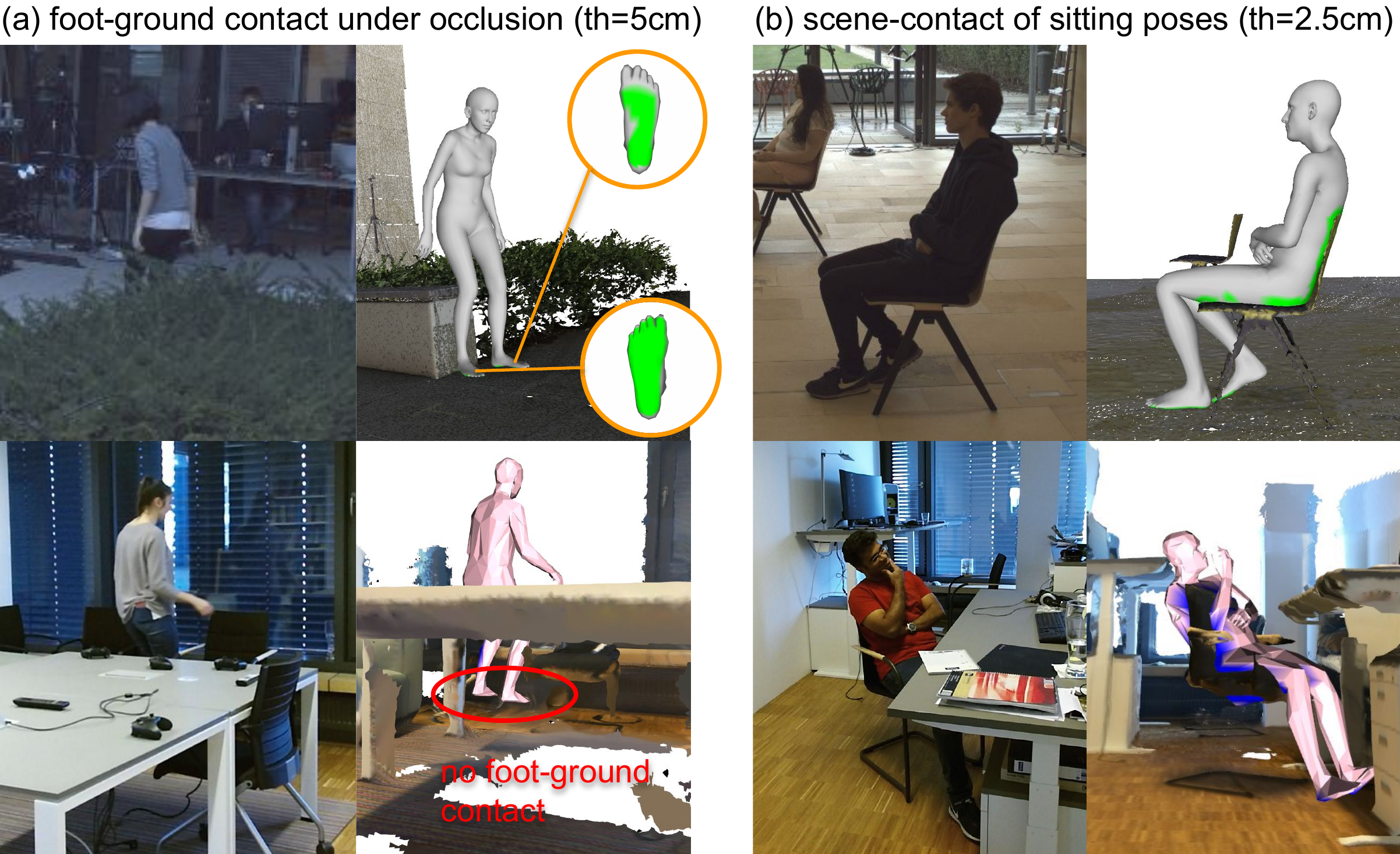}
  \caption{Comparison of \hsc annotations in \datasetname (top) and POSA\cite{Hassan:CVPR:2021} (bottom). The noisy body fits in PROX \cite{PROX:2019} result in undesirable \hsc labels in POSA: (a) no foot-ground contact under occlusion; (b) severe penetration with chairs.}
  \label{fig:richvsprox}
  \vspace{-1em}
\end{figure}

We place the reconstructed bodies into pre-scanned \threeD scenes to estimate the body-scene contact. 
The scene mesh and HDR textures were acquired using an industrial laser scanner, Leica RTC360.
To put the bodies in the scene, we solve the rigid transformation between camera coordinates and scan coordinates with manually identified correspondences.
To annotate human-scene contact automatically, our approach is similar to POSA \cite{Hassan:CVPR:2021}.
Specifically, for each vertex on the body mesh, we compute the point-to-surface distance to the scene scan. 
If the distance is lower than a threshold and the normal is compatible, we accept the hypothesis that it is \emph{in contact}.
Considering the thickness of shoe soles, the threshold is 5cm for the vertices at the bottom of feet and 2.5cm for the rest of body.
This is different from POSA, which uses 5cm for the whole body to collect training data from PROX \cite{PROX:2019}.
Furthermore, the pseudo-ground-truth body poses in PROX are obtained by fitting the SMPL-X template to monocular RGBD data.
As shown in the bottom row of Fig.~\ref{fig:richvsprox}, PROX accuracy suffers from occlusion, sometimes resulting in severe penetration with the scene.
The errors in body fits are carried over to the ground-truth \hsc data for POSA.
In contrast, in \datasetname, bodies are recovered from \multiv data, which reduces the issues caused by occlusion and depth ambiguity.


\section{Methods: \methodname}

Here we introduce \methodname for dense \hsc estimation from a single image.
This relies on \datasetname, described in Sec.~\ref{sub-sect:dataset} in detail.
%
Existing \hsc methods usually take a multi-stage approach. 
Given an input image, they first reconstruct the body mesh and use it as a proxy to infer contact.
Formally, let $f$ denote the function recovering a body mesh $M$ from the input image $I$, $M=f(I)$.
$f$ can be an energy-minimization process such as \cite{SMPL-X:2019} or a neural network as in \cite{kanazawa2018end,Kocabas_PARE_2021}.
To estimate contact, \sota methods differ from each other in two ways: 
(1) the features extracted from $M$, \eg, Euclidean distance to the \threeD scene, velocity and body poses (cf.~Table \ref{table:relatedwork}); 
(2) the prediction functions, \eg, simple thresholding, neural network, or physics engine.
With a slight abuse of notation, we denote these feature extraction and contact estimation processes collectively as $g$, 
which takes the body $M$ as input and predicts a contact vector $\contactvector = g\left(  M\right) $.
Each element in $\contactvector$ is $1$ if the corresponding part of the body (vertex, joint or body part) is in contact with the scene, and $0$ otherwise.
For example, $g$ represents the decoder of a conditional VAE in POSA \cite{Hassan:CVPR:2021}, taking the vertex locations of $M$ as input,
while in \cite{rempe2021humor,RempeContactDynamics2020,Shimada2020PhysCap}, $g$ is a MLP operating on the motion of $M$.

\begin{figure}[t]
  \centering
  \includegraphics[width=1\linewidth]{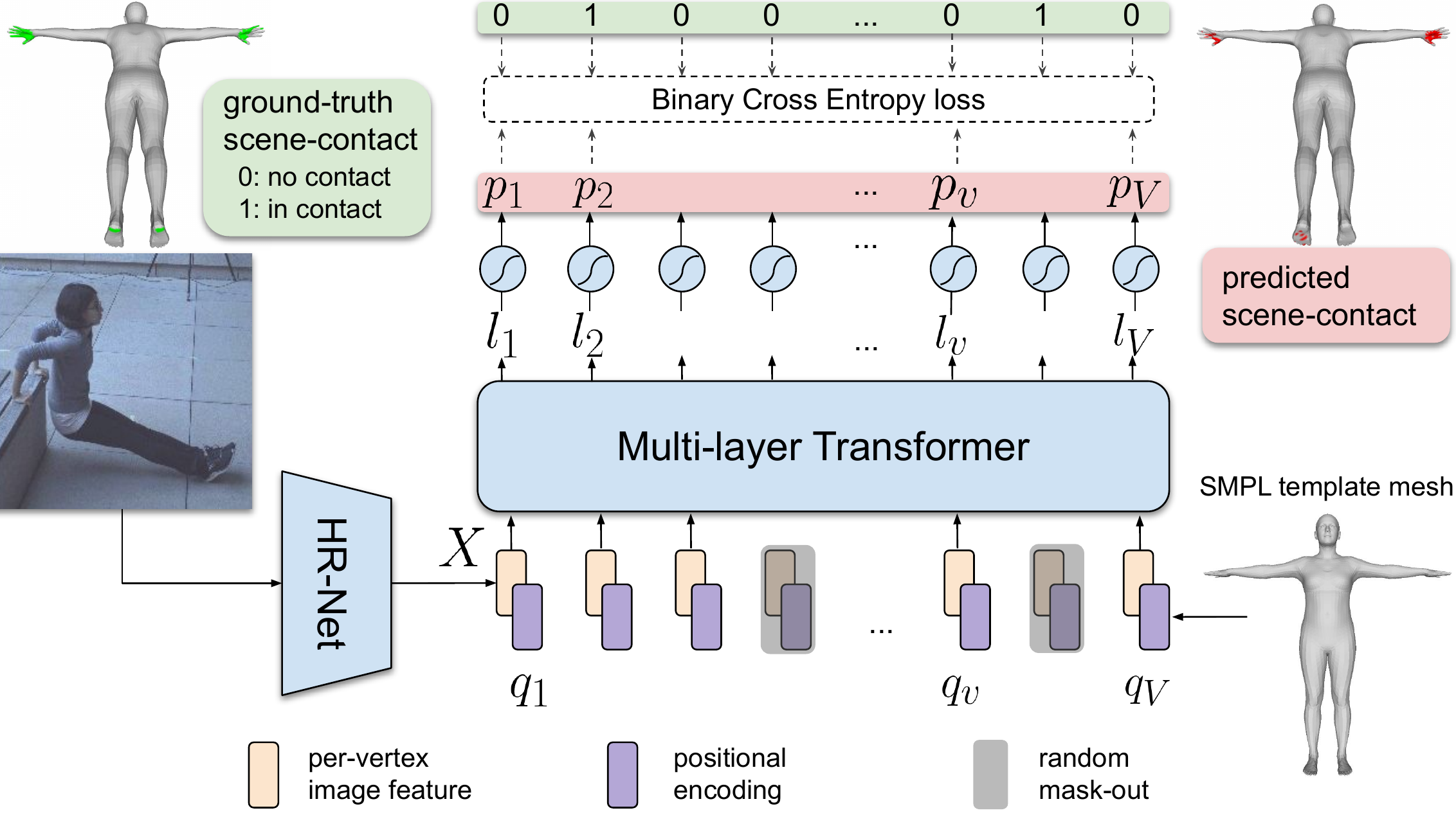}
  \caption{\textbf{{\methodname} model architecture.} Given an input image, \methodname predicts dense per-vertex contact labels by exploiting image information, without reconstructing 3D poses or 3D bodies. 
  }
  \label{fig:bstro}
  \vspace{-1em}
\end{figure}

With this formulation, the body-scene contact $\contactvector$, 
whether defined on a dense mesh or on a set of sparse joints/parts, 
is a composite function of $g$ and $f$: $\contactvector=g \circ f(I)$, 
where $g$ is agnostic 
to the input image. 
In contrast, our goal is to detect dense body-scene contact directly from the input $I$: $\contactvector=g(I)$.
To our knowledge, this was explored only for self-contact \cite{Fieraru_2021_AAAI} and person-person contact \cite{Fieraru_2020_CVPR} and only at a coarse region level, not the vertex level.

We use SMPL as the body representation for \methodname, 
hence $\contactvector \in \{0,1\}^\numVert$, where $\numVert$=6890 is the number of vertices on a SMPL mesh, as opposed to $V$=10475 on a SMPL-X mesh.
The reason for this choice is that a SMPL-X mesh has nearly 50\% of the vertices on the head, which rarely participates in natural body-scene contact, so we would like to reduce the dimensionality of the output space. See \supmat~for more discussion of this design choice.


We model $g$ as a neural network and train it end-to-end in a supervised way with the $(I,\contactvector)$ pairs sampled from \datasetname. 
The network architecture is designed based on our key observation.
That is, regions in contact are not directly observable due to occlusion.
However, there is rich information in the image to tell which parts of the body are in contact with the scene.
Estimating \hsc from images is therefore inherently a ``hallucination'' task.
Without really ``seeing'' the regions in contact, the network needs to explore the image freely and attend to regions it finds informative. 

We use a multi-layer transformer \cite{devlin2018bert} to learn such a non-local relationship from data
and propose the \emph{Body-Scene contact TRansfOrmer} (\methodname). 
Figure~\ref{fig:bstro} visualizes the architecture of \methodname.
It takes an image of a person as input, 
extracts features $X \in \mathbb{R}^{2048}$ with a CNN backbone, and appends vertex locations of the SMPL template as positional encoding. 
The feature after concatenation is denoted as $q \in \mathbb{R}^{2051}$.
The input query of the transformer is a set of $q$: $Q = \{q_v\}_{v=1}^{V}$.
The transformer outputs an array of logits $l_v$, which, after applying 
sigmoid functions, result in elements $p_v\in [0,1]$ encoding the probability of vertex $v$ being in contact. 
Finally, the dense scene-contact vector $\contactvector$ is obtained by thresholding  $p_v$ at 0.5.
Note that \methodname is a non-parametric method, in spirit similar to \cite{lin2021end-to-end} that makes prediction for each vertex directly without passing through a parametric model.
\begin{figure*}[t]
  \centering
  \includegraphics[width=1\linewidth]{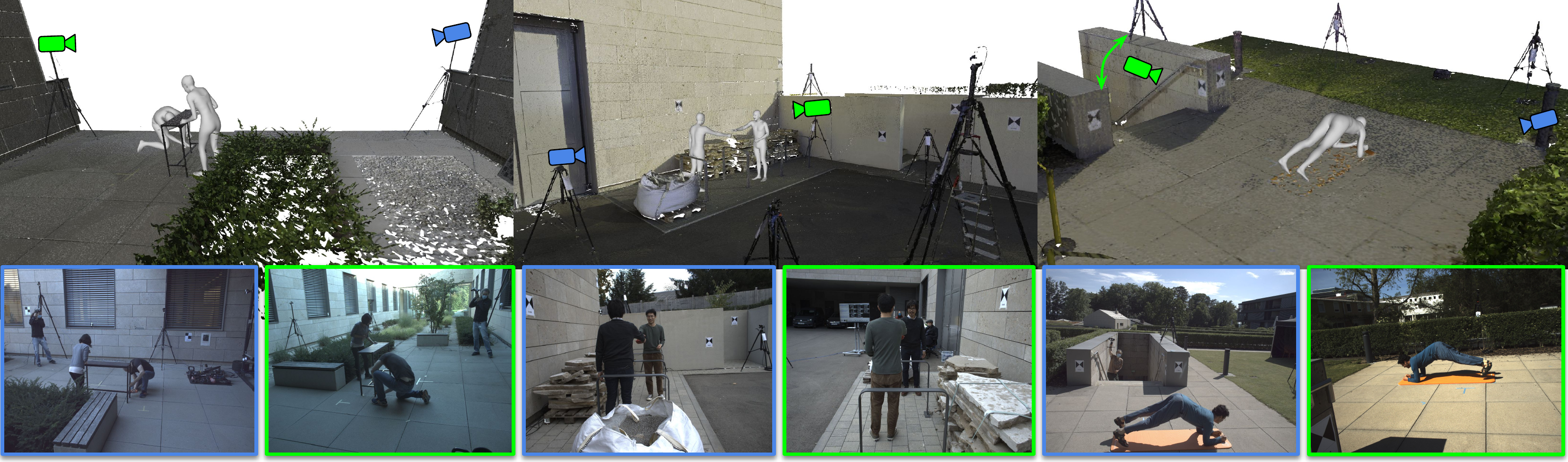}
  \caption{\textbf{{\datasetname} dataset.} In each scene we capture subjects' motions with 6-8 static cameras and, for some scenes, with 1 additional moving camera. Top row: scans of three example outdoor scenes with example \threeD body meshes. Bottom row: RGB images from these scenes. The color border matches the camera icon of the same color. 
  }
  \vspace{-1em}
  \label{fig:rich}
\end{figure*}

\noindent \textbf{Training.}
We apply the binary cross entropy loss between the ground truth contact and the predicted contact probability $p_v$.
One can think of this as a multi-label classification problem, 
where each category (vertex) has its own probability of being true (in contact) or not.

To gain robustness to occlusion, 
we employ Masked Vertex Modeling (MVM) 
\cite{lin2021end-to-end}.
Specifically, at each iteration, we randomly mask out some queries in $Q$ 
and still ask the transformer to estimate contact for all vertices.
In order to predict the output of a missing query, 
the model has to explore other relevant queries. 
This simulates occlusions where bodies are only partially visible and also encourages the network to hallucinate contact.

\section{\datasetname Dataset} \label{sub-sect:dataset}

We capture \numsub subjects performing various human-scene interactions in \numscene static 3D scenes with 6-8 static cameras and, in some scenes, with an additional (untracked) moving camera (Fig.~\ref{fig:rich} rightmost scene).
\textcolor{black}{Subjects gave prior written informed consent for the capture, use, and  distribution of their data for research purposes. 
The experimental methodology has been reviewed by the University of T\"{u}bingen Ethics Committee with no objections.}

\datasetname has in total \nummvvideo single or multi-person \multiv videos, with a total of \numpose posed \threeD body meshes, together with \numpose dense full-body contact labels in both SMPL-X and SMPL mesh topology, and \numimage high resolution (4K) images.
Compared to PROX, \datasetname consists of mostly outdoor environments with areas of roughly 60m\textsuperscript{2}.
The images in \datasetname are real, not limited to a single subject, have dynamic backgrounds and varied viewpoints. 
All these features make it suitable for training and evaluating monocular \hsc methods.
Figure \ref{fig:rich} shows several examples of \datasetname.

In addition, since \datasetname provides SMPL-X fits, \ie, pseudo-ground-truth human poses and shapes, 
it can also serve as a monocular or \multiv \hps benchmark. 
It contains more subjects than 3DPW \cite{vonMarcard20183dpw}, more accurate body shapes than AGORA \cite{Patel:CVPR:2021}, and real human-scene interaction unlike Human3.6M \cite{h36m_pami}. 
In our experiments we analyze the performance of \sota \hps methods with respect to body-scene contact. 
Such analyses are not feasible with existing \hps datasets.
\begin{figure*}[t]
  \centering
  \includegraphics[trim=000mm 1mm 000mm 000mm, clip=true, width=1\linewidth]{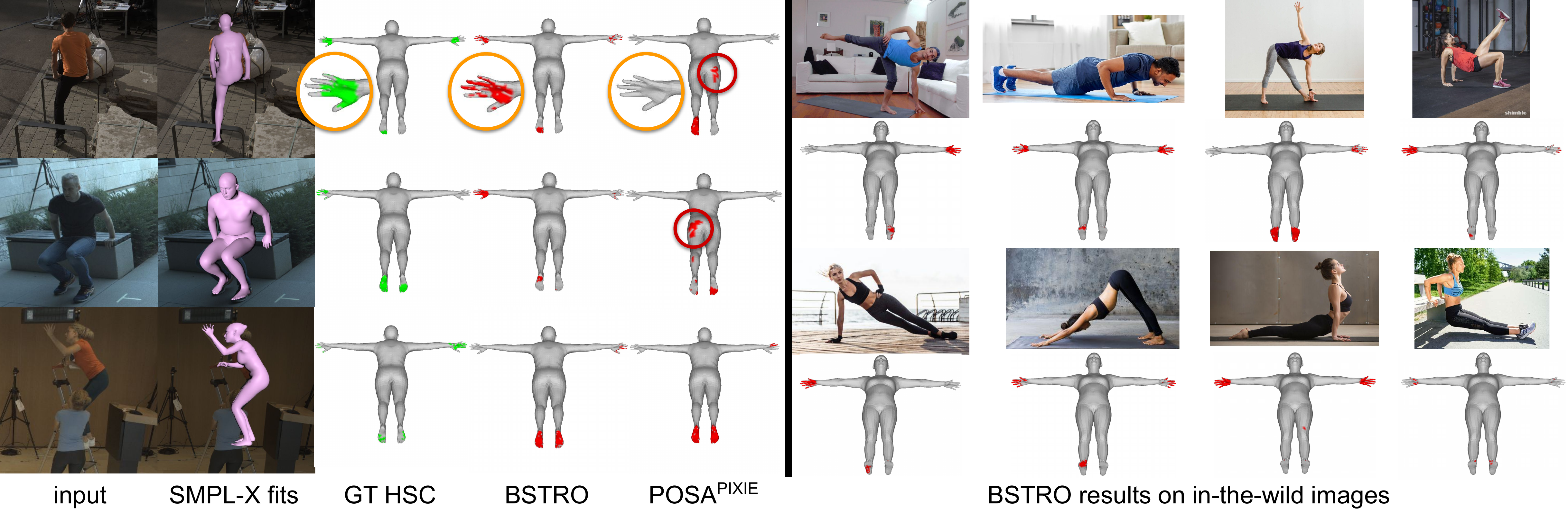}
  \caption{Left: qualitative results on \datasetname -test. GT \hsc stands for ground-truth human-scene contact computed from the SMPL-X fits and scene scans.
  \methodname estimates more accurate scene contact than POSA\textsuperscript{PIXIE}. Right: qualitative results on in-the-wild images. 
  }
  \label{fig:qualitative}
 \vspace{-1em}
\end{figure*}
\section{Experiments}
\label{experiments}
\subsection{Dataset Split}
We split \nummvvideo multiview videos in \datasetname into \nummvvideoTr, \nummvvideoVal, \nummvvideoTe for training, validation, and testing purposes, respectively.
The test set consists of several subsets designed for varied evaluation protocols.
Each subset is defined by whether or not each of three attributes 
has been observed in training:
scene, human-scene interaction, and subject.
The most challenging subset is when they are all unseen in \datasetname-train.
The split ensures there is one completely withheld scene and 7 unseen subjects in the test set. 
See \supmat~for more breakdowns in terms of 3D bodies and images. 

\subsection{Evaluation Metrics and Baselines}

We apply standard detection metrics (precision, recall, and F1 score) to evaluate the estimated dense \hsc.
Since vertex density varies over the SMPL template, the
same number of false positives, say, on the palm and on the thigh 
correspond to different areas on the body surface, 
but this is not reflected in the scores above. 
To better understand how well an \hsc method estimates contact, 
we additionally consider a measure that translates the count-based scores to errors in metric space. 
Specifically, 
for each vertex predicted in contact, 
we compute its shortest \emph{geodesic distance} to a ground-truth vertex in contact. 
If it is a true positive, this distance is zero; 
if not, this distances indicates the amount of prediction error along the body.

We evaluate three \hsc baselines on the \datasetname-test.
Zou \etal~\cite{zou2020reducing} use the velocity of 4 2D keypoints on the feet to predict contact;
\humor \cite{rempe2021humor} estimates contact for 8 joints while reconstructing human motions.
These two methods estimate contact for sparse joints, not dense vertices,
so we mark all vertices that correspond to a joint as contact when the method predicts the joint is in contact.
POSA \cite{Hassan:CVPR:2021} requires a 3D body mesh in the canonical space as input to sample dense body contact.
We consider two choices of 3D bodies for POSA: (1) using the results from a \sota body regressor PIXIE \cite{feng2012pixie},
or (2) using ground-truth bodies to evaluate the impact of errors in estimated body pose.

\subsection{Main Results}
The results on \datasetname-test are reported in Table \ref{table:quan_ex_sota}.
We see that \humor yields overall lowest detection scores and  highest geodesic errors.
This is partially due to the fact that it only considers contact with an even ground plane, 
while \datasetname-test contains more varied real scene interactions. 

POSA, in general, has higher recall compared to other methods.
This, however, comes with a cost of precision, meaning that there are many false positives.
Comparing rows (c) and (d) we see that recall is significantly better when using ground-truth bodies.
\methodname yields significantly better precision but with lower recall than POSA. 
Still, it has the highest F1 score and lowest geodesic error, which shows that it strikes a good balance between precision and recall.
Figure \ref{fig:qualitative} shows some visual examples. 
\datasetname has accurately fitted SMPL-X bodies and body-scene contact.
Given an input image, \methodname estimates scene contact that is closer to the ground truth, 
whereas POSA\textsuperscript{PIXIE} yields false positives frequently (red circles) and sometimes misses the contact on the hands.
While the training dataset is limited, \methodname also works on in-the-wild images,
as shown in the right part of Fig.~\ref{fig:qualitative}.

\subsection{Generalization}
To analyze how well \methodname generalizes, 
we split \datasetname-test into several subsets. 
Each subset represents whether \methodname has observed similar images of the three attributes: scene, human-scene interaction (\hsi), and subject.
This allows us to inspect the importance of each attribute, and to know which aspect future methods should focus on.
Note that this is a unique feature of \datasetname, as existing \hsc datasets from \mocap~\cite{mixamo,AMASS:2019} and \hps datasets  \cite{h36m_pami,vonMarcard20183dpw,joo2021eft} do not support such an analysis.


In Table \ref{table:quan_ex_analysis}, \cmark~means \methodname has seen similar images of that attribute during training, while \xmark~means it has not.
For example, images in row (a) share the same subjects and similar \hsi with training data but the scenes are new. 
Intuitively, this is an easy subset and indeed the scores
are high in this scenario.
Once \hsi is withheld, the performance drops (row (e)).
This drop is more pronounced 
than the drop caused by withholding a subject (row (c)). 
Comparing each of the rows (c,d,e) to row (f), 
we observe that seeing similar \hsi at training helps the most.
Seeing the same scenes or same subjects does not guarantee gains in performance. 
Finally, row (f) represents the most challenging subset, where scene, \hsi, and subjects are all unseen during training.
We see that \methodname still yields results that are comparable to other subsets. 
Subset (c) contains many images with person-person occlusion, \eg, Fig.~\ref{fig:qualitative} bottom left, which partially explains why it is the most challenging.

\subsection{\hps Evaluation on \datasetname -test}
Besides evaluating human-scene contact, \datasetname can also serve as a benchmark for monocular \hps methods.
Unlike existing \hps benchmarks with real images such as 3DPW \cite{vonMarcard20183dpw} or Human3.6M \cite{h36m_pami}, 
the \emph{real} scene contact in \datasetname enables a new way of analyzing the performance of an \hps method.
In particular, we use PIXIE \cite{feng2012pixie}, a recent monocular \hps method, to regress SMPL-X bodies from  \datasetname-test.
We compare the estimated SMPL-X bodies with the pseudo-ground-truth SMPL-X fits from Sec.~\ref{sub-sect:mvinit}, 
and compare the error when body-scene contact is present or absent.

We consider Mean Per-Joint Position Error (MPJPE) and Vertex-to-Vertex Error (V2V) to measure the discrepancies in joints and body meshes respectively.
For freely moving cameras, we apply Procrustes alignment (PA) before calculating the two errors, hence PA-MPJPE and PA-V2V.
Procrustes alignment factors out differences in rotation, scale and translation, focusing on measuring the difference in ``pure body poses.''
PA hides many sources of errors so we use it only when ground-truth camera extrinsic parameters are not available.
For calibrated cameras, on the other hand, 
we factor out only translation by aligning the estimated and ground-truth bodies to their pelvis locations, denoted with a prefix ``TR.''
We ignore foot-ground contact, which is ubiquitous, and compare the results when there is meaningful scene contact vs.~no scene contact.

On average, images containing meaningful scene contact yield 214.0mm/172.81mm TR-MPJPE/TR-V2V, 
higher than 161.81mm/121.71mm for images with no contact other than foot-ground contact.
This is partially due to the fact that scene contact usually comes with scene occlusion, 
and this shows a direction where monocular \hps methods can improve.
The corresponding errors in moving cameras are 84.15mm/83.16mm PA-MPJPE/PA-V2V for images with meaningful contact and 63.67mm/64.37mm for those without.
We again observe that the presence of scene contact makes \hps more challenging, yielding higher errors.
This shows that scene contact impacts all aspects of the problem: from pure body poses to global orientation and translation.

    

\begin{table}[t]
\centering
\footnotesize
\begin{tabular}{l|cccr}
\toprule
     Methods & precision $\uparrow$ & recall $\uparrow$  & F1 $\uparrow$ & geo.~error $\downarrow$ \\ \hline 
     a.~Zou \etal~\cite{zou2020reducing}                        & 0.277      &  0.609  & 0.359  &  17.48cm\\ \hline 
     b.~\humor \cite{rempe2021humor}                            & 0.248       & 0.527  &0.314  & 25.35cm \\ \hline 
     c.~POSA \cite{Hassan:CVPR:2021}\textsuperscript{GT}        & 0.375      &  0.768 & 0.464 &  19.96cm \\ \hline 
     d.~POSA \cite{Hassan:CVPR:2021}\textsuperscript{PIXIE}     & 0.312      &  0.699  & 0.399 & 21.16cm  \\ \hline 
     \rowcolor{lightgray}
     e.~\methodname & \textbf{0.699} &  \textbf{0.774}  & \textbf{0.708}   & \textbf{10.98cm}\\ 
     \bottomrule
    
\end{tabular}
\caption{\textbf{Evaluation on \datasetname -test.} POSA\textsuperscript{GT} means taking ground-truth bodies as input, while POSA\textsuperscript{PIXIE} takes the estimated bodies from PIXIE\cite{feng2012pixie}.}
\vspace{-1em}
\label{table:quan_ex_sota}
\end{table}

\begin{table}[t]
\centering
\footnotesize
\begin{tabular}{lccc|cccr}
\toprule
     & scene & \hsi & subject & p. $\uparrow$ & r. $\uparrow$   & F1 $\uparrow$ & geo.~err.~$\downarrow$\\ \hline 
    a.& \xmark & \cmark & \cmark   & \textbf{0.766} &  0.819 & \emph{0.769}   & \textbf{4.77cm} \\ \hline 
    b.& \cmark & \cmark & \xmark   & 0.734 &  0.756 & 0.718   & 9.64cm\\ \hline
    c.& \cmark & \xmark & \xmark   & 0.556 &  0.593 & 0.520   & 25.11cm\\ \hline 
    d.& \xmark & \cmark & \xmark   & \emph{0.753} &  \textbf{0.870} & \textbf{0.790}   & \emph{6.61cm}\\ \hline 
    e.& \xmark & \xmark & \cmark   & 0.644 &  \emph{0.820} & 0.705    &16.64cm\\ \hline 
    f.& \xmark & \xmark & \xmark   & 0.682 &  0.814 & 0.721    &9.00cm\\ \hline
    g.& \multicolumn{3}{c}{full \datasetname-test set} & 0.699 &  0.774  & 0.708   & 10.98cm\\ 
     \bottomrule
    
\end{tabular}
\caption{The performance of \methodname on each subset of \datasetname-test. p.: precision; r.: recall. \cmark/\xmark: observed attribute at training. 
Bold/italic: the best and the 2\textsuperscript{nd} best results in each metric.
}
\label{table:quan_ex_analysis}
\end{table}

\section{Conclusion}
\label{conclusion}

While there is rapid progress on estimating 3D human pose and shape from images, much of this work ignores the scene and the interaction of the body with that scene.
Capture and analysis of body-scene contact, however, is critical to understanding human action in detail.
To address this, and to help the 
research community
study this problem, we created \datasetname, a new dataset with challenging natural video sequences, high-resolution \threeD scene scans, ground-truth body shapes,  high-quality reference poses, and detailed \threeD contact labels.
We use the contact information to train a new method (\methodname) that takes a single image of a person interacting with a scene and infers the \threeD contacts on their body.
We also use the dataset to evaluate human pose estimation and find that scenes with significant contact cause problems for the state of the art.
The dataset and code are available  for research purposes.

\noindent \textbf{Limitations and future work}. 
\datasetname considers only contact with static scenes 
so does not account for the body contact with dynamic scenes, \eg, 
with hand-held objects, or human-human interaction.
One extension would estimate the rigid-body pose of an object given its 3D model and simultaneously reconstruct the hand/body that interacts with it.
Another interesting direction would jointly estimate the body pose, shape, and scene contact in one single network.

\noindent\textbf{Acknowledgments}.
We thank Taylor McConnell, Claudia Gallatz, Mustafa Ekinci, Camilo Mendoza, Galina Henz, Tobias Bauch, and Mason Landry for the data collection and cleaning.
We thank all participants who contributed to the dataset, Paola Forte for PIXIE experiment and 	
Benjamin Pellkofer for IT support.
Daniel Scharstein was supported by NSF grant IIS-1718376.

{\small
\noindent\textbf{Disclosure:}
\href{https://files.is.tue.mpg.de/black/CoI_CVPR_2022.txt}{https://files.is.tue.mpg.de/black/{CoI\_CVPR\_2022.txt}}
}
{\small
\bibliographystyle{ieee_fullname}
\bibliography{main}
}

\newpage
	\appendix
	{\noindent\Large\textbf{Supplementary Material}}
	\newline

\begin{figure*}
  \centering
  \includegraphics[trim=000mm 000mm 000mm 000mm, clip=false, width=1.00 \linewidth]{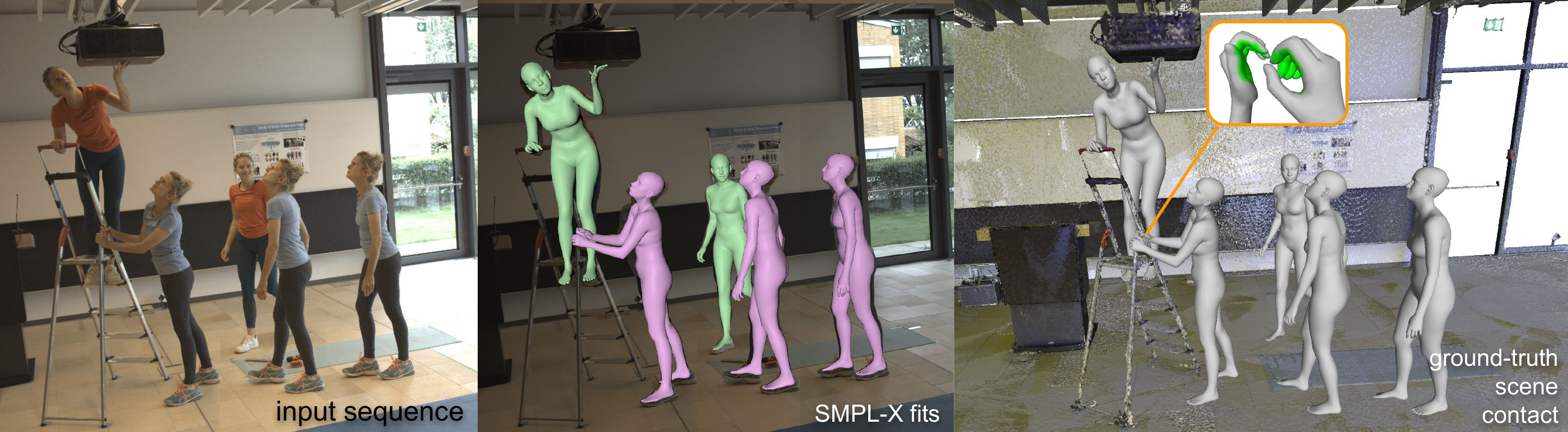}
  \caption{The \textbf{\datasetname dataset} contains multiple people interacting with a real scene.
    It provides complex natural images, precise 3D scene scans, pseudo ground-truth \smplX bodies, and dense body contact labels.
  }
  \label{fig:suppteaser}
\end{figure*}
			
	\section{SMPL-X vs.~SMPL \hsc labels}
\begin{figure}[t]
  \centering
  \includegraphics[width=1\linewidth]{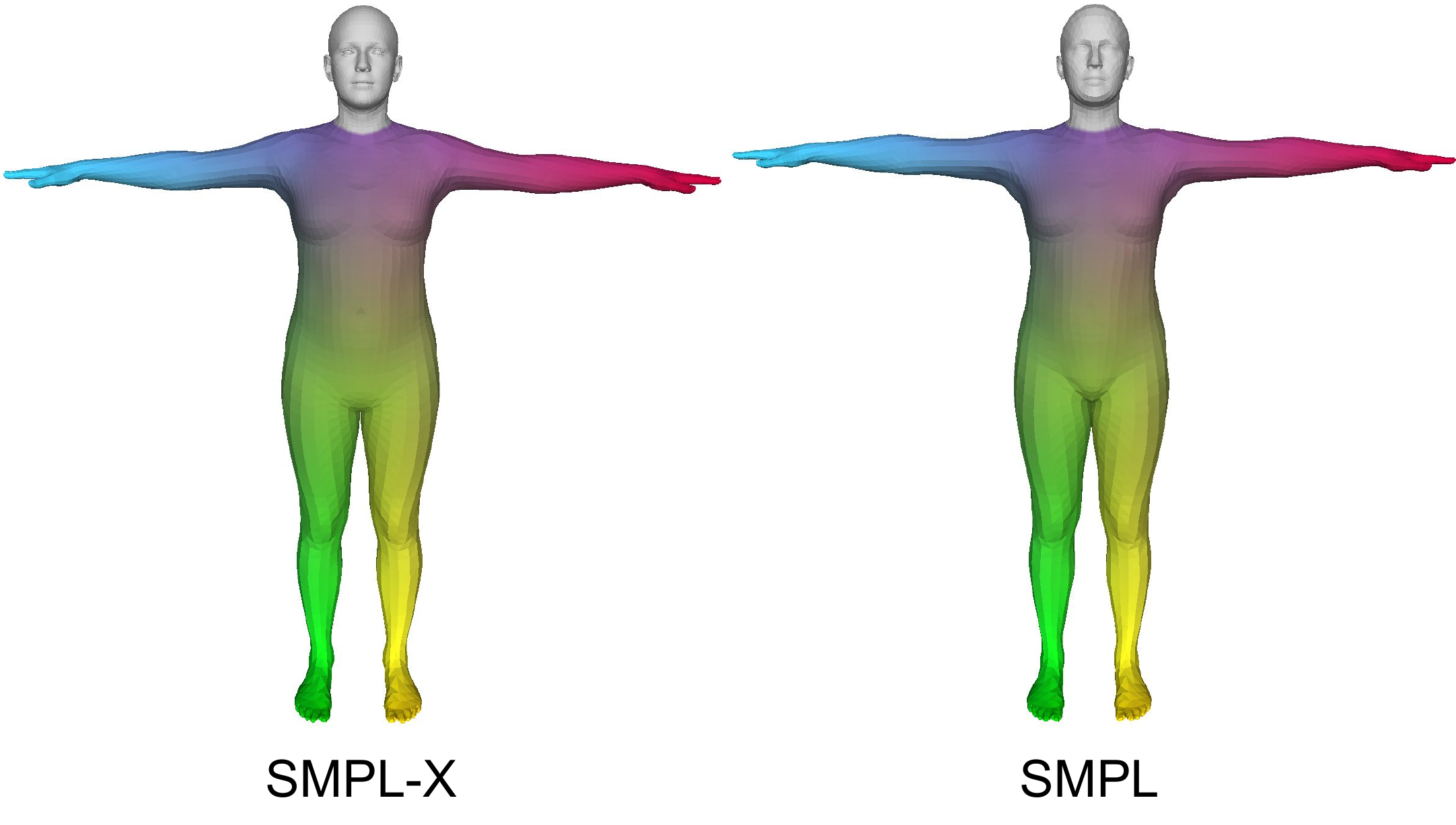}
  \caption{ SMPL-X and SMPL bodies share the same set of vertices for regions below the neck. The same vertices are visualized in the same colors.
  }
  \label{fig:smplsmplxcorres}
\end{figure}

We build \datasetname by fitting a SMPL-X template to multi-view data and compute the human-scene contact (\hsc) as explained in the Sec.~\ref{methods} and Sec.~\ref{sub-sect:dataset} of the main paper (Fig.~\ref{fig:suppteaser}).
The contact labels are defined in \smplX format and we map them to \smpl format for training \methodname.
This is feasible since there is an 1-to-1 correspondence between \smplX and \smpl vertices below the neck, as shown in Fig.~\ref{fig:smplsmplxcorres}.

With this mapping, we convert the ground-truth \hsc labels from \smplX to \smpl without losing information.
As a result, we benefit from realistic hand articulation in \smplX and still keep the dimension of the output space small (\smpl). 
Such a mapping also makes \datasetname a suitable \hsc benchmark for both body models. 
Since the two models share the set of vertices of interest, choosing either of them does not influence the detection scores or errors.

On the other hand, the human pose and shape (\hps) parameters of the two models differ.
Converting \hps parameters between \smplX and \smpl requires extra processing \cite{model_transfer} and one always loses the hand articulation when converting from \smplX to \smpl.
Therefore, \datasetname provides only \smplX as pseudo ground truth. 
To evaluate methods that regress \smpl parameters using \datasetname, users should convert \smpl to \smplX, which does not result in a loss of information.

\section{\datasetname Dataset}
The \nummvvideo  multi-view videos in \datasetname are recorded at a rate of 30 frames per second. 
We separate them into subsets of \nummvvideoTr, \nummvvideoVal, \nummvvideoTe for
training, validation, and testing purposes, respectively.
This amounts to 303K, 149K, 125K images of 4K resolutions (in total \numimage), 
and 40K, 18K, 32K 3D \smplX bodies along with dense scene-contact labels (in total \numimage)
in each subset.
By ``body" here, we mean any \smplX mesh. Note that the number of unique ``people" in the dataset is much smaller than the number of bodies because every posed mesh constitutes a separate body.

Compared to the recent \hps dataset AGORA \cite{Patel:CVPR:2021}, \datasetname has more \threeD bodies (\numpose vs.~4K), more images (\numimage vs.~19K) and more accurate body shapes (registrations to minimally-closed scans \cite{Hirshberg:ECCV:2012} vs.~clothed scans \cite{zhang2017detailed}).
It has more subjects in varied body shapes than 3DPW \cite{vonMarcard20183dpw} (22 vs.~18) and subjects are in natural clothing as opposed to those in Human3.6M \cite{h36m_pami}.
Last but not least, \datasetname provides high-quality scene scans and scene contact labels that none of the above datasets provides. 
\section{Bone-orientation Term $E_O$}
Following the illustration in Fig.~\ref{fig:bl_mvconsensus}(a) of the main paper, 
the bone-orientation term $E_O$ factors out the residual of the parent joint $\epsilon_1$ from the residual of the child joint $\epsilon_2$: 
\begin{align}
\nonumber
    r_2    &= \epsilon_2 - \epsilon_1, \\
    \nonumber
    &= (j'_2-j_2) - (j'_1-j_1), \\
    \nonumber
    &= (j'_2-j'_1) - (j_2-j_1), \\
    \nonumber
    &= b'_2 - b_2,
\end{align}
where $b'_2=j'_2-j'_1$ and $b_2=j_2-j_1$ denote the ``bone vector'' of target points (detected landmarks) and estimated \smplX joints respectively.
It follows that
\begin{align}
    \|r_2\|_2^2 = \| b'_2\|_2^2 +  \| b_2\|_2^2 -  b_2 ^\top  b'_2.
\end{align}
Since $b'_2$ involves only the detected landmarks and $\|b_2\|$ is fixed given a constant body shape $\beta$, 
the first two terms are constant when optimizing the multi-view objective $E_\text{mv}$.
$\|r_2\|_2^2 $ is therefore minimized when $b_2 ^\top  b'_2$ is maximized, \ie, when $b_2$ has the same orientation as $b'_2$.


\section{\methodname Implementation Details}
We sample \datasetname-train to build the image-\hsc pairs $(I, \contactvector)$ for training \methodname.
For each sequence, we consider only every other frame, and for each frame, we use the dynamic view and one randomly selected static view, or two static views if no moving camera is available.
This sampling strategy ensures sufficient variations in viewpoints and background, while keeping the total number of the training pairs tractable. 

We train with in total 24K $(I, \contactvector)$ pairs from \datasetname-train and use Adam \cite{adam} optimizer with an initial learning rate of 1e-4 for 100 epochs.
The HR-Net backbone is initialized with the weights pre-trained on ImageNet \cite{deng2009imagenet}, Human3.6M \cite{h36m_pami} or 3DPW \cite{vonMarcard20183dpw}. 
The best checkpoint is selected by the best performance on \datasetname-validation with \datasetname-test completely withheld.
We refer interested readers to \cite{lin2021end-to-end} for the architecture of the multi-layer transformer.


\end{document}